\documentclass[final]{cvpr}

\usepackage{times}
\usepackage{epsfig}
\usepackage{graphicx}
\usepackage{amsmath}
\usepackage{amssymb}

\usepackage{soul}
\usepackage{bm}
\usepackage{enumitem}
\usepackage{subfigure}
\usepackage{url}
\usepackage{algorithm}
\usepackage{algorithmic}
\usepackage{braket}
\usepackage{array}

\usepackage[pagebackref=true,breaklinks=true,colorlinks,bookmarks=false]{hyperref}

\def\cvprPaperID{1962} 
\def\confYear{CVPR 2021}

\newcommand{\bc}{\mathbf{c}}
\newcommand{\bu}{\mathbf{u}}
\newcommand{\br}{\mathbf{r}}
\newcommand{\bx}{\mathbf{x}}

\newcommand{\bR}{\mathbf{R}}
\newcommand{\bK}{\mathbf{K}}

\newcommand{\vel}{\bm{\omega}}

\newcommand{\be}{\mathbf{e}}
\newcommand{\cE}{\mathcal{E}} 
\newcommand{\cO}{\mathcal{O}} 
\newcommand{\cF}{\mathcal{F}}
\newcommand{\cD}{\mathcal{D}}
\newcommand{\cI}{\mathcal{I}}
\newcommand{\cL}{\mathcal{L}}
\newcommand{\cG}{\mathcal{G}}
\newcommand{\cM}{\mathcal{M}}
\newcommand{\cN}{\mathcal{N}}
\newcommand{\bM}{\mathbf{M}}
\newcommand{\cT}{\mathcal{T}}

\newcommand{\cW}{\mathcal{W}} 

\newcommand{\bv}{\mathbf{v}}

\newcommand{\cS}{\mathcal{S}}

\newcommand{\bzero}{\mathbf{0}}
\newcommand{\btheta}{\bm{\theta}}
\newcommand{\bI}{\mathbf{I}}

\newtheorem{lem}{Lemma}

\newtheorem{cor}{Corollary}

\newtheorem{defn}{Definition}

\DeclareMathOperator*{\argmin}{arg\,min}

\begin{document}

\title{Spatiotemporal Registration for Event-based Visual Odometry}

\author{Daqi Liu~~~~~~~Álvaro Parra~~~~~~~Tat-Jun Chin\\
		School of Computer Science, The University of Adelaide\\
		{\tt\small \{daqi.liu, alvaro.parrabustos, tat-jun.chin\}@adelaide.edu.au}
	}

\maketitle

\begin{abstract}
A useful application of event sensing is visual odometry, especially in settings that require high-temporal resolution. The state-of-the-art method of contrast maximisation recovers the motion from a batch of events by maximising the contrast of the image of warped events. However, the cost scales with image resolution and the temporal resolution can be limited by the need for large batch sizes to yield sufficient structure in the contrast image\footnote{See supplementary material for demonstration program.}. In this work, we propose spatiotemporal registration as a compelling technique for event-based rotational motion estimation. We theoretically justify the approach and establish its fundamental and practical advantages over contrast maximisation. In particular, spatiotemporal registration also produces feature tracks as a by-product, which directly supports an efficient visual odometry pipeline with graph-based optimisation for motion averaging. The simplicity of our visual odometry pipeline allows it to process more than 1 M events/second. We also contribute a new event dataset for visual odometry, where motion sequences with large velocity variations were acquired using a high-precision robot arm\footnote{Dataset: \url{https://github.com/liudaqikk/RobotEvt}}.
\end{abstract}

\section{Introduction}

Due to their ability to asynchronously detect intensity changes, event sensors are well suited for conducting visual odometry (VO) in applications that require high temporal resolution~\cite{delbruck2013robotic,mueggler2014event,weikersdorfer2014event,mueggler2017fast,vidal2018ultimate}, \eg, high-agility robotic manipulation, fast manoeuvring aerial vehicles. However, to fully reap the benefits of event sensing for VO, efficient algorithms are required to process event streams with low-latency to accurately recover the experienced motion.

\begin{figure}[ht]\centering
\subfigure{\includegraphics[width=0.49\columnwidth]{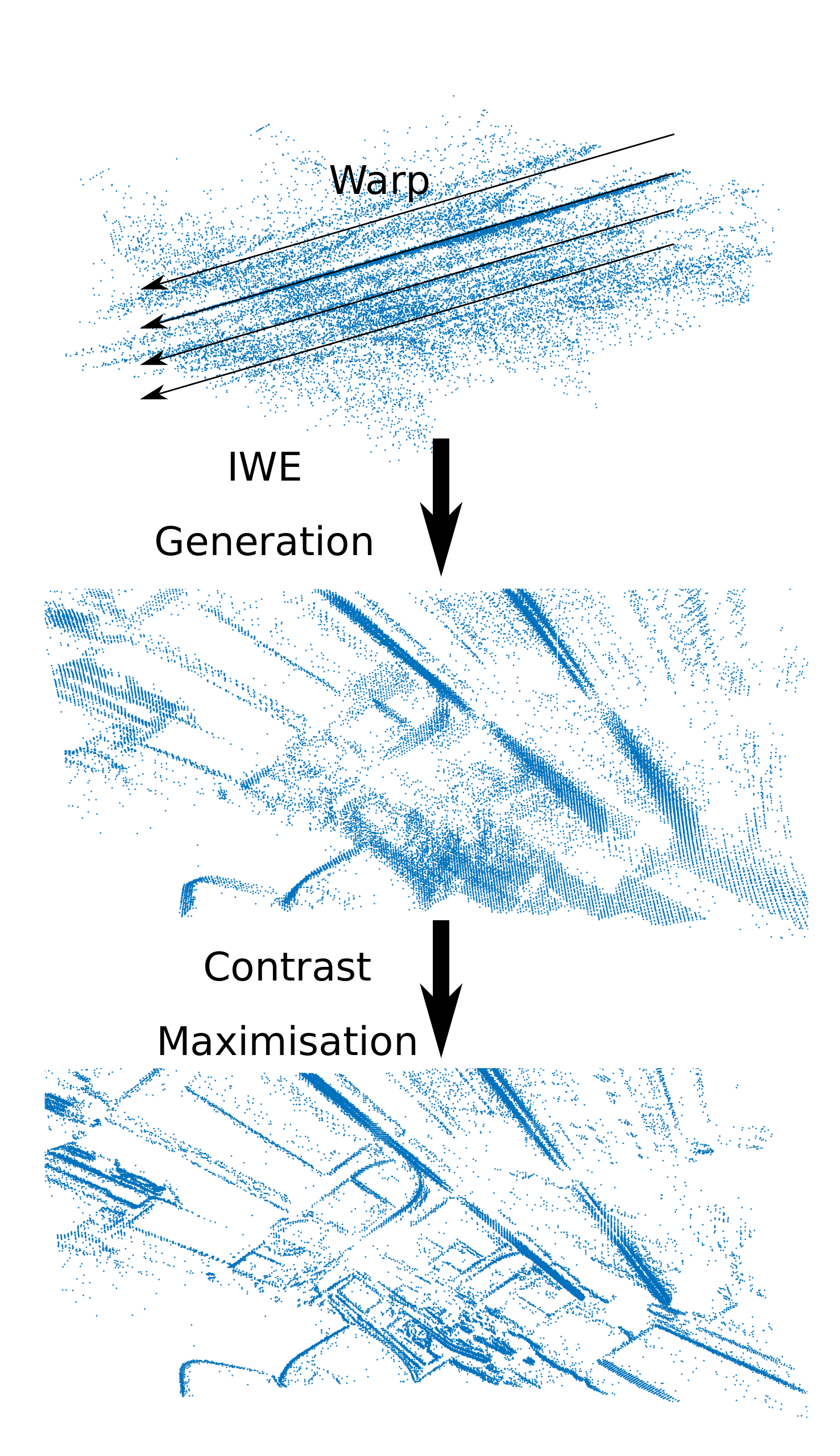}}
\subfigure{\includegraphics[width=0.49\columnwidth]{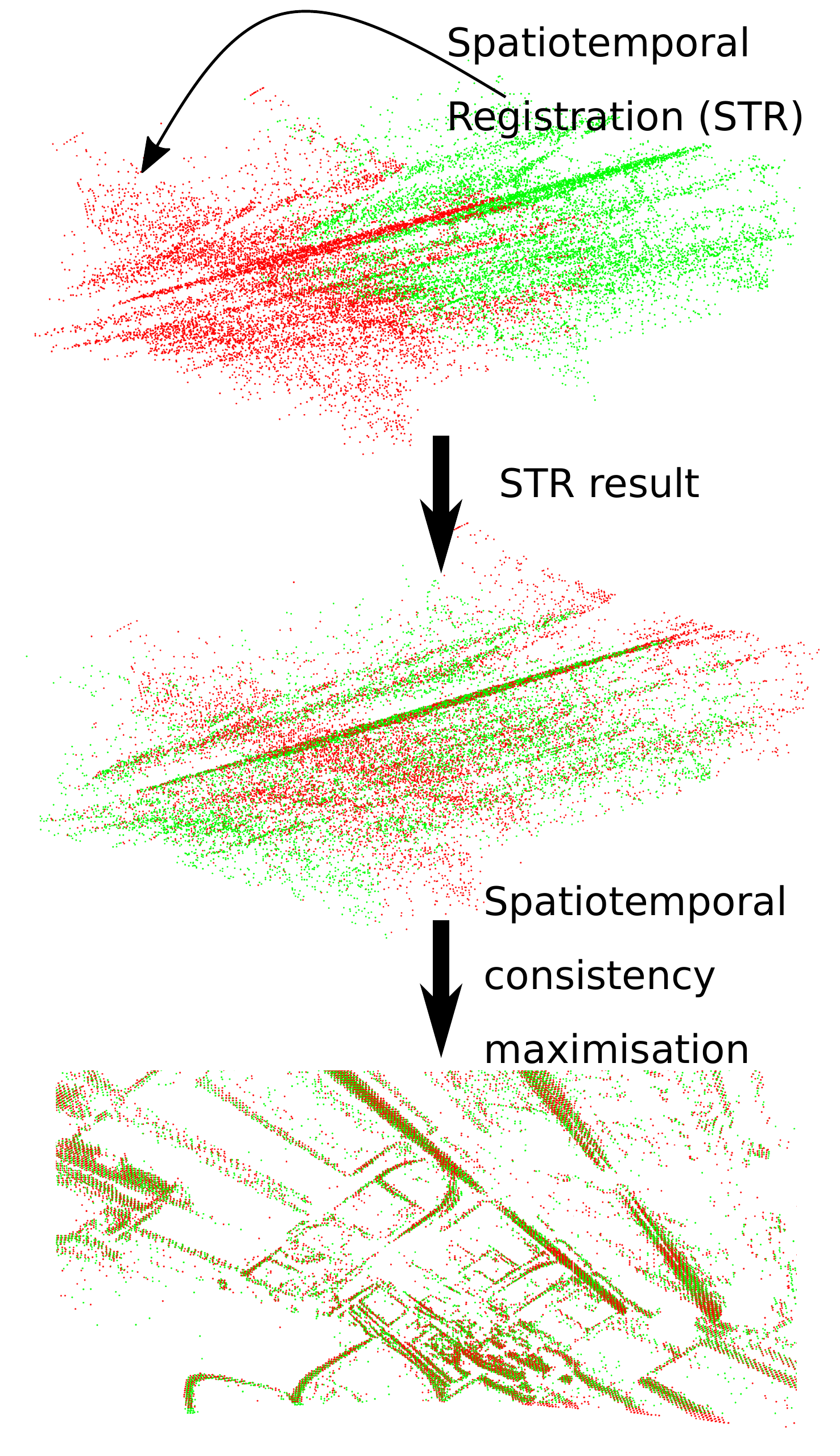}}
\caption{Conceptual difference between contrast maximisation and spatiotemporal registration for event-based motion estimation.}
\label{fig:concept}
\end{figure}

An event sensor produces an event stream $\cS = \{ \be \}$, where each $\be = ( \bu, t, p )$ is a tuple containing the 2D image coordinates $\bu$, time stamp $t$ and polarity $p$ associated with a brightness change that exceeded the preset threshold. In scenarios where the event camera (i.e., event sensor plus optics and other components) moves in a static environment, the events are triggered mainly by the camera motion. The goal of VO is to recover the camera motion from $\cS$.



Many event-based VO methods~\cite{zhou2020event,rebecq2016evo,gallego2017accurate,rebecq2017real} conduct ``batching", where small subsets of $\cS$ are processed incrementally. Each batch $\cE = \{ \be_i \}^{N}_{i=1} \subset \cS$ is acquired over a time window $\cT = [\alpha, \beta]$, where each $\be_i = (\bu_i,t_i,p_i)$ is associated with a 3D point in the camera FOV that triggered $\be_i$ at time $t_i \in \cT$. The core task is to estimate the relative motion $\cM$ between $\alpha$ and $\beta$ from $\cE$. The estimated $\cM$ is then subject to the broader VO pipeline (more in Sec.~\ref{sec:related}).
\subsection{Contrast maximisation}\label{sec:cm}

A state-of-the-art approach to estimate $\cM$ from $\cE$ is contrast maximisation (CM)~\cite{gallego2018unifying}. Parametrising $\cM$ by a vector $\vel \in \Omega$ and letting $\cD = \{\bx_j\}_{j=1}^{P}$ be the image domain (the set of pixel coordinates) of the event sensor, each candidate $\vel$ yields the image of warped events (IWE)
\begin{equation}\label{eq:eventimg}
H(\bx_j;\vel) = \sum_{i=1}^N\kappa_\delta(\bx_j-f(\bu_i, t_i ;\, \vel)),
\end{equation}
where $f$ warps $\bu_i$ to a position in $H$ by reversing the motion $\vel$ from $t_i$ to the start of $\cT$. The form of $f$ depends on the type of motion $\cM$(see~\cite{gallego2018unifying} for details). The warped events are aggregated by a kernel $\kappa_\delta$ with bandwidth $\delta$, \eg,
\begin{equation}
\kappa_\delta(\bx) = \exp(\|\bx\|_2/2\delta^2).
\end{equation}
The contrast of $H$ is given by
\begin{equation}\label{eq:contrast}
C(\vel) =\dfrac{1}{P}\sum_{j=1}^P (H(\bx_j;\vel)-\mu(\vel))^2,
\end{equation}
where $\mu(\vel)$ is the mean intensity of $H$ the image.
Both $C$ and $\mu$ are functions of $\vel$ since $H$ is dependent on $\vel$.  CM estimates $\vel$ by maximising $C(\vel)$, the intuition being that the correct $\vel$ will yield a sharp image $H$; see Fig.~\ref{fig:concept}.

Previous studies found CM effective in a number of event-based VO tasks~\cite{gallego2018unifying}, especially where $\cM$ is a rotation, i.e., $\Omega = SO(3)$. However, there are a couple of fundamental weaknesses in CM, as described in the following.

\vspace{-1em}
\paragraph{Computational cost}

Maximising $C(\vel)$ can be done using conjugate gradient~\cite{gallego2018unifying} and branch-and-bound~\cite{Liu_2020_CVPR}. Note that the cost to compute~\eqref{eq:contrast} depends on both
\begin{itemize}[topsep=0.25em,itemsep=0.25em,parsep=0em]
\item the number of pixels $P$; and
\item the number of events $N$ in the batch $\cE$.
\end{itemize}
While $P$ is a constant of the event sensor, $N$ depends on the motion speed and scene complexity. A higher $P$ increases the FOV and hence tends to increase $N$, however, the cost of $C(\vel)$ will increase with $P$ even if $N$ is constant.

\begin{figure}[ht]\centering
	\subfigure[Original scene.]{\includegraphics[width=0.49\columnwidth]{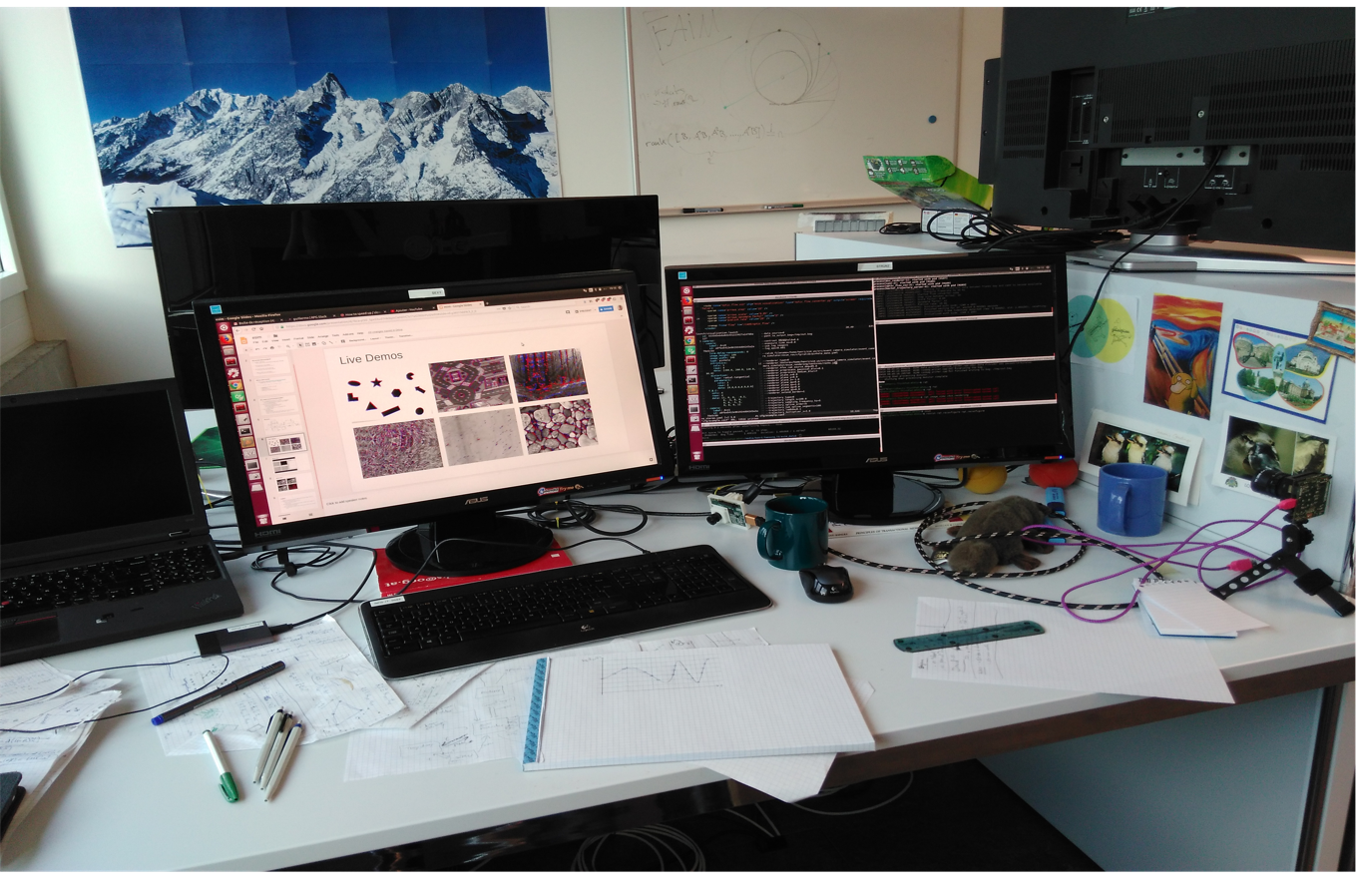}}
	\subfigure[Runtime vs num.~of pixels $P$.]{\includegraphics[width=0.49\columnwidth]{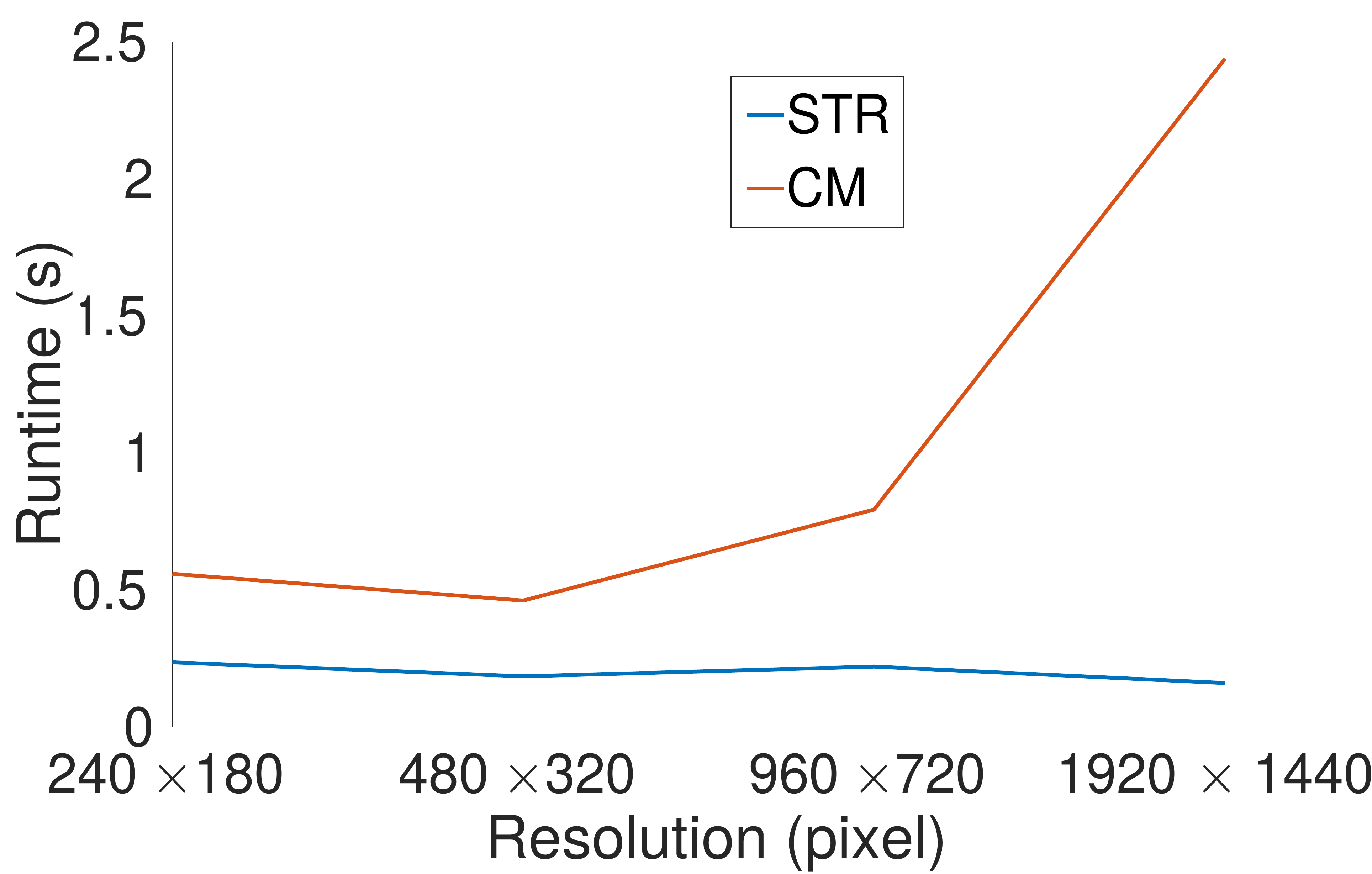}}
	\caption{Using the image in panel (a) as input to ESIM~\cite{rebecq2018esim}, we generated synthetic event batches as outputs of event sensors of varying resolution, from $P = 240 \times 180$ to $P = 1920 \times 1440$ pixels. By tuning the duration $|\cT|$, the batch size $N$ was fixed at $15,000$. Panel (b) illustrates the average runtime of CM and STR on the generated data as a function of resolution.}
	\label{fig:runtime}
\end{figure}

The basic analysis above indicates that the cost of CM (regardless of the algorithm) will also scale with both $P$ and $N$. Fig.~\ref{fig:runtime} plots the runtime of CM (using conjugate gradient) on input instances with increasing $P$ and constant $N$, which shows a clear uptrend. While early event sensors have low resolutions (\eg, $240 \times 180$ on iniVation Davis 240C), current sensors can have up to 1 Megapixels (\eg, $1280 \times 720$ on Prophesee 720P CD, $1280 \times 800$ on  CeleX-V). Following industry trends, event sensors will likely continue to increase in resolution. To maintain the efficiency of CM on high-resolution sensors, a separate heuristic to reduce the ``image resolution" is needed (note that this is different sparsifying $\cE$ by reducing the number of events $N$).

\begin{figure}[ht]\centering
\subfigure[Error vs batch duration $|\cT|$.]{\includegraphics[width=0.49\columnwidth]{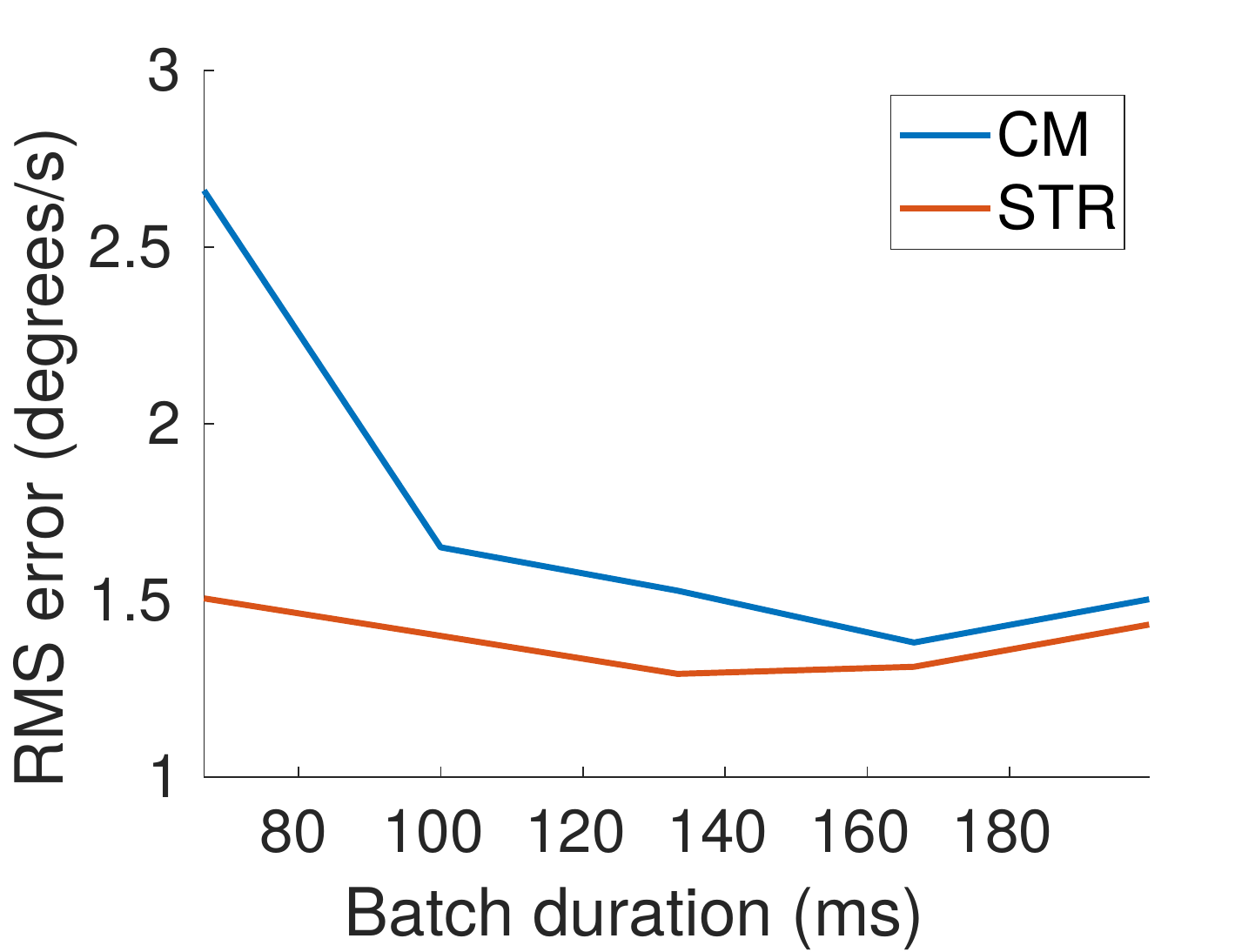}}
\subfigure[Runtime vs batch size $N$.]{\includegraphics[width=0.49\columnwidth]{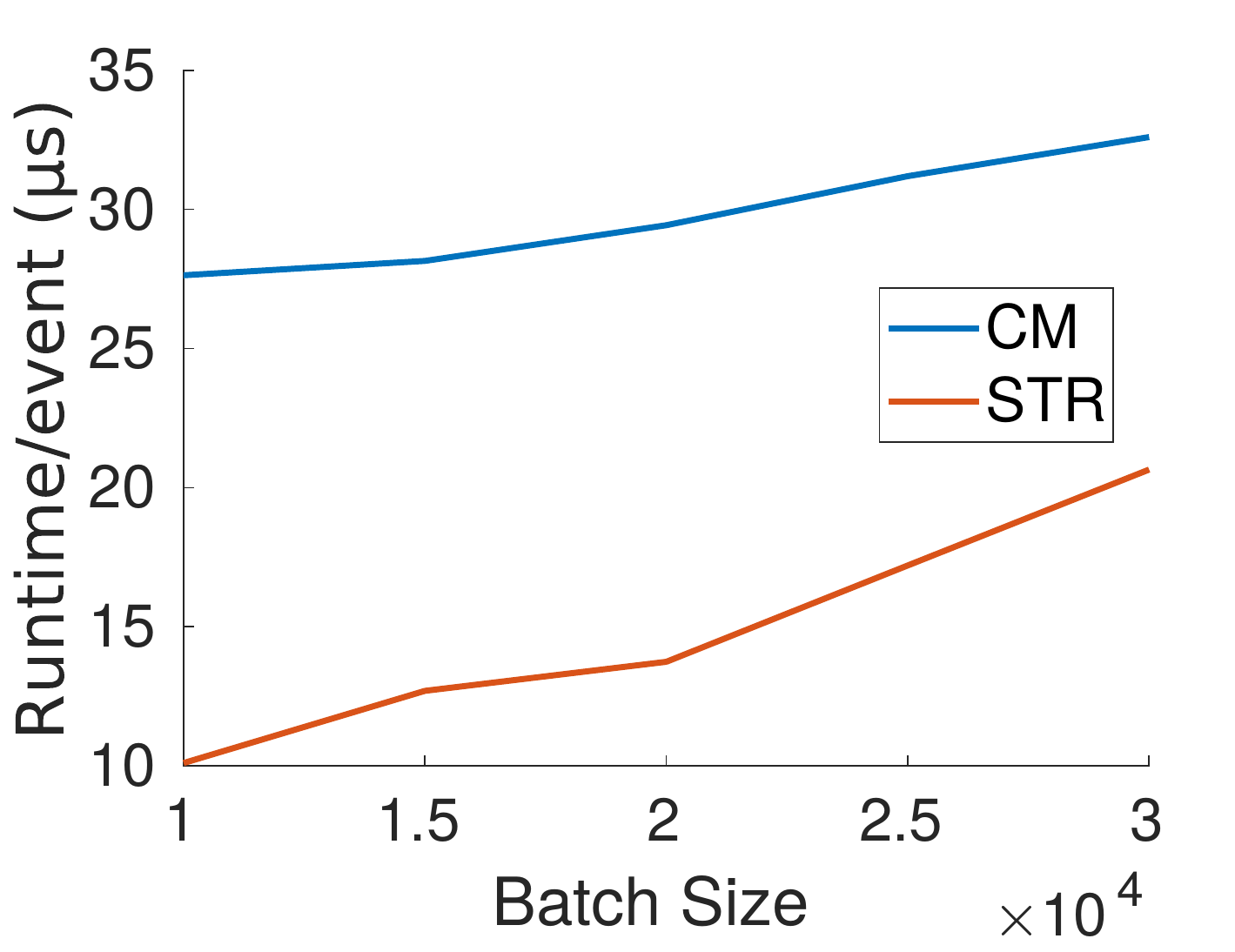}}
\caption{Motion estimation error and runtime of CM and STR.}
\label{fig:lowspeed}
\end{figure}

\vspace{-1em}
\paragraph{Temporal resolution}

Intuitively the accuracy of estimating $\cM$ depends on capturing sufficient ``structure" in $\cE$. For a fixed scene and motion rate, the amount of structure in $\cE$ increases with the duration $|\cT| = \beta - \alpha$~\cite{mueggler2015lifetime}. Conversely to achieve VO with high temporal resolution, $|\cT|$ should be as small as possible to minimise batching effects. The conflicting demands indicate a maximum temporal resolution achievable by an event-based motion estimation technique.

Fig.~\ref{fig:lowspeed} shows the motion estimation accuracy of CM on batches $\cE$ of different durations $|\cT|$ from sequence \texttt{PureRot\_Mid\_Off} of our event dataset (Sec.~\ref{sec:dataset}). Unlike the experiment in Fig.~\ref{fig:runtime}, the number of events $N$ were varied according to $|\cT|$. Note the degradation in accuracy as $|\cT|$ decreases (i.e., $N$ decreases), which indicates a lower temporal resolution of CM (more results in Sec.~\ref{sec:results}).

\subsection{Our contributions}

We propose spatiotemporal registration (STR) as a cogent alternative to CM; see Fig.~\ref{fig:concept} on the concept of STR. Despite the relative simplicity of STR, it has not been thoroughly investigated for event-based motion estimation. Specifically, we will justify STR by examining the conditions in which it is valid (Sec.~\ref{sec:str}) and demonstrate that it is generally as accurate for rotational motion estimation but does not suffer from the fundamental weaknesses of CM demonstrated in Sec.~\ref{sec:cm} (more results in Sec.~\ref{sec:results}).

Further, unlike CM, STR produces feature correspondences as a by-product (see Fig.~\ref{fig:concept}). This directly enables a novel event-based VO pipeline (Sec.~\ref{sec:evo}) that conducts feature tracking and motion averaging. To support our experiments, we build a new event dataset for VO using a high-precision robot arm (Sec.~\ref{sec:dataset} and our dateset website).

\section{Related works}\label{sec:related}

The CM framework has been improved in several directions. Stoffregen \etal~\cite{Stoffregen_2019_CVPR} adjusted the contrast objective by introducing ``sparsity" to improve the accuracy of motion estimation. They later integrated CM into the segmentation task~\cite{Stoffregen_2019_ICCV}. Globally optimal CM was proposed in~\cite{Liu_2020_CVPR,pengglobally}, where~\cite{Liu_2020_CVPR} further accelerated the algorithm by integer quadratic programming relaxation. Seok and Lim~\cite{Seok_2020_WACV} dropped the constant velocity assumption and replaced linear interpolation by Bezier curve. In general, the improvements above tend to increase cost, which discourage real-time applications.

Closer to our work is Nunes and Demiris~\cite{Nunes_2020_ECCV} who proposed an entropy minimisation framework (EM) for event-based motion estimation. Their approach maximises the similarity (minimising the entropy) between the feature vector of events. Like our proposed STR method, EM also obviates the need to compute the image of warped events. However, although a truncated kernel was used to accelerate their algorithm, EM is still too expensive for online application, as results in Sec.~\ref{sec:results} will show.



The techniques surveyed above can be considered ``direct methods" since all events are utilised in the computation. Unlike direct methods, ``feature-based" methods achieve VO by detecting and tracking simple structures in the event data, such as circles~\cite{ni2012asynchronous} and lines~\cite{conradt2009pencil}. To handle the more complex scene, traditional frame-based feature detectors are utilised on motion-compensated event images~\cite{rebecq2017real,vidal2018ultimate}, frames~\cite{tedaldi2016feature,kueng2016low} and time surfaces (TS)~\cite{vasco2016fast,mueggler2017fast,alzugaray2018asynchronous} recently. A TS~\cite{gallego2019event} is a reconstructed image that each pixel records the temporal information of the last event as “intensity” of the image. Based on the detected features, Alzugarary et al.~\cite{alzugaray2018ace} propose a descriptor and a tracker that employed the descriptors for event data. Zhu et al.~\cite{zhu2017event} present a feature tracking based on Expectation Maximisation (EM). They later propose visual-inertial odometry (VIO)~\cite{zihao2017event} system by fusing IMU and their feature trackers. On the other hand, ~\cite{rebecq2017real} utilises Kanade–Lucas–Tomasi feature tracker~\cite{baker2004lucas} on the motion-compensated event images with the motion from IMU. Note that all the feature-based methods highly rely on a different heuristic for keypoint detection and tracking, which can quickly lose track without IMU. 

Learning-based method has gained more attention recently, and there have been several works that solve the event-based VO with unsupervised learning~\cite{zhu2019unsupervised,ye2018unsupervised}, and spiking network~\cite{gehrig2020event}. Both~\cite{zhu2019unsupervised} and ~\cite{ye2018unsupervised} follow the framework and architecture from SfMLearner~\cite{zhou2017unsupervised} and propose some changes for event-based setting. However, Since lack of training data, all learning-based methods cannot be generalised to different environments, which overfitting to the training data. Moreover, ~\cite{zhan2020visual,bian2020unsupervised} show the limitations in pure rotation motion of learning-based method.


\section{Spatiotemporal registration}\label{sec:str}

In this section, we describe the proposed event-based relative motion estimation technique, including its fundamental underpinnings and optimisation algorithm.


\subsection{Motion model}


For a batch of events $\cE$ acquired over time duration $\cT = [\alpha, \beta]$, we represent using
\begin{align}
\bM_t = \left[ \begin{matrix} \bR_t & \bc_t \\ \bzero & 1 \end{matrix} \right]
\end{align}
the absolute pose of the event camera at time $t \in \cT$, where $\bR_t \in SO(3)$ and $\bc_t \in \mathbb{R}^3$ are respectively the absolute orientation and position of the camera at the same time  $t$. Let $a$ and $b$ be two time instances in $\cT$, where
\begin{align}
\alpha \le a \le b \le \beta.
\end{align}
The relative motion between $a$ and $b$ is given by
\begin{align}\label{eq:relmotion}
\bM_{a,b} &= \bM_{b}\bM_{a}^{-1}\nonumber\\
&= \left[ \begin{matrix}  \bR_{b}\bR_{a}^T  & -\bR_{b}\bR_{a}^T\bc_{a} + \bc_{b}  \\ \bzero & 1 \end{matrix} \right].
\end{align}

We follow many previous works~\cite{Liu_2020_CVPR,gallego2017accurate,pengglobally,gehrig2020event} to focus on \emph{rotational odometry}, which is useful for a number of applications, e.g., video stabilisation~\cite{gallego2018unifying}, panorama construction~\cite{kim2016real}, star tracking~\cite{chin2019star,bagchi2020event}. This allows to assume pure rotational motion for $\bM_t$, where $\bc_t = \bzero$ for all $t$ and the relative motion~\eqref{eq:relmotion} reduces to the relative rotation
\begin{align}
\bM_{a,b} &= \left[ \begin{matrix}  \bR_{b}\bR_{a}^T  &  \bzero  \\ \bzero & 1 \end{matrix} \right].
\end{align}
More succinctly, the relative rotation between $a$ and $b$ is
\begin{align}\label{eq:relrot}
\bR_{a,b} &:= \bR_b\bR_a^T,
\end{align}
and setting $a = \alpha$ and $b = \beta$ yields $\bR_{\alpha,\beta}$, which is the target relative motion $\cM$ to be estimated from $\cE$.

The short duration of $\cE$ (e.g., in the $ms$ range) further motivates to assume constant angular velocity in the period $\cT$. Specifically, the absolute orientation can be written as
\begin{align}\label{eq:constangvel}
\bR_t = \exp(\left[ t \vel \right]_\times)\exp(\left[ \btheta_0 \right]_\times).
\end{align}
for all $t \in \cT$, where $\exp$ is the exponential map.
In more detail, vector $\vel \in \mathbb{R}^3$ defines the angular velocity in period $\cT$, where the direction $\hat{\vel}$ of $\vel$ provides the axis of rotation and the length $\| \vel \|_2$ of $\br$ specifies the rate of change of the angle of the rotation about $\hat{\vel}$. The initial orientation at time $\alpha$ is given by $\btheta_0$, and $t \vel$ is the rotational increment on $\btheta_0$ from time $a$ to time $t$.

Applying the BCH formula~\cite{bch} on~\eqref{eq:relrot} yields
\begin{align}\label{eq:motionmodel}
\bR_{a,b} &= \exp(\left[ b\vel \right]_\times)\exp(\left[ -a\vel \right]_\times) \nonumber\\
&= \exp(\left[(b - a)\vel\right]_\times),
\end{align}
where terms involving $\btheta_0$ cancel out, and $\left[ a\vel \right]_\times$ and $\left[ b\vel \right]_\times$ commute and hence the Lie bracket $[\left[ a\vel \right]_\times,\left[ b\vel \right]_\times] = 0$. The significance of this derivation is encapsulated in the following lemma.

\begin{lem}\label{lem}
Assuming that the camera undergoes pure rotational motion with constant angular velocity~\eqref{eq:constangvel} in the period $\cT$, the relative rotation $\bR_{a,b}$ between any $a, b \in \cT$ with $a \le b$ depends only on the difference $b - a$ and $\vel$.
\end{lem}

\begin{figure}[ht]\centering
\subfigure{\includegraphics[width=0.8\columnwidth]{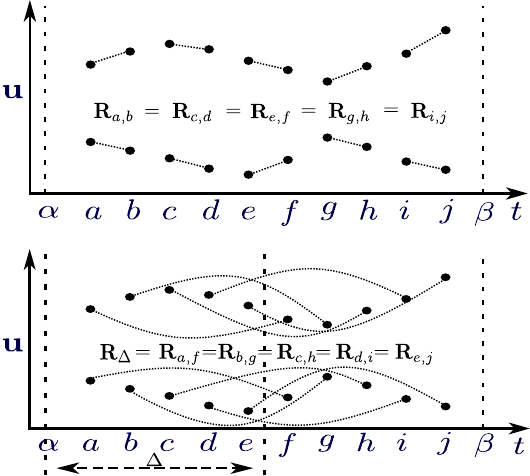}}
\caption{Equivalent relative rotations (see Corollary~\ref{cor}) and their spatiotemporally consistent event correspondences (see Definition~\ref{def2}). In the bottom example, the relative rotations are also equivalent to $\bR_{\Delta}$, where $\Delta = 0.5(\beta - \alpha)$.}
\label{fig:str}
\end{figure}

A straightforward corollary of Lemma~\ref{lem} is as follows, which is also illustrated in Fig.~\ref{fig:str}.

\begin{cor}\label{cor}
Under the motion model assumed in Lemma~\ref{lem}, $\bR_{a,b} = \bR_{c,d}$ for all time instances $a,b,c,d$ in the period $\cT$ such that $a \le b$, $c \le d$ and $b-a = d-c$.
\end{cor}

%
%


\subsection{Event-based relative motion estimation}

We exploit the insights above to estimate $\cM = \bR_{\alpha,\beta}$ from event batch $\cE$. First, we define the notion of spatiotemporal consistency and event correspondences.

\begin{defn}[Spatiotemporal consistency]\label{def}
Under the motion model assumed in Lemma~\ref{lem}, a relative rotation $\bR_{a,b}$, with $a, b \in \cT$ and $a \le b$, and a pair of events $\be = (\bu, t, p)$ and $\be^\prime = (\bu^\prime, t^\prime, p^\prime)$, where $t, t^\prime \in \cT$ and $t \le t^\prime$, are spatiotemporally consistent if
\begin{itemize}[topsep=0.25em,itemsep=0.25em,parsep=0em]
\item $t^\prime - t = b - a$ (temporal consistency); and
\item $\hat{\bu}^\prime = \bR_{a,b}\hat{\bu}$ (geometric consistency),
\end{itemize}
where $\hat{\bu}$ is the backprojected ray (a unit vector)
\begin{align}
\hat{\bu} = \frac{\bK^{(1:2)} \tilde{\bu}}{\bK^{(3)} \tilde{\bu}}
\end{align}
of the image point $\bu$, where $\tilde{\bu} = [\bu^T, 1]^T$ and $\bK^{(1:2)}$ and $\bK^{(3)}$ are respectively the first-2 rows and 3rd row of the camera intrinsic matrix $\bK \in \mathbb{R}^{3\times 3}$~\cite{Glover2017b} (similarly for $\bu^\prime$).
\end{defn}

\begin{defn}[Event correspondence]\label{def2}
An event correspondence $\langle \be, \be^\prime \rangle$ are a pair of two events $\be$ and $\be^\prime$ that are spatiotemporally consistent with a relative rotation.
\end{defn}

Intuitively, an event correspondence is associated with the same 3D scene point that was observed during $\cT$. Fig.~\ref{fig:str} also shows valid event correspondences. In particular, Fig.~\ref{fig:str} depicts event correspondences for the relative rotation 
\begin{align}
\label{eq:RDelta}
\bR_{\Delta} := \bR_{\alpha, \alpha+(\beta - \alpha)/2},
\end{align}
where to simplify notation we also define
\begin{align}
\Delta = (\beta - \alpha)/2.
\end{align}

We approach the estimation of $\bR_{\alpha,\beta}$ by recovering $\bR_{\Delta}$ from the noisy event batch $\cE = \{ \be_i \}^{N}_{i=1} = \{(\bu_i, t_i, p_i)\}^{N}_{i=1}$  acquired over period $\cT = [\alpha,\beta]$. To this end, we first separate $\cE$ into two mutually exclusive subsets
\begin{align}
\cE_\alpha &= \{ \be_i \in \cE \ \mid \alpha \le t_i \le \alpha + \Delta \},\\
\cE_\beta &= \{ \be_i \in \cE \mid \alpha + \Delta < t_i \le \beta \}.
\end{align}
Note that since the events in $\cE$ are ordered in time by construction, we can write
\begin{align}
\cE_\alpha &= \{ \be_1, \be_2, \dots, \be_M \},\\
\cE_\beta &= \{ \be_{M+1}, \be_{M+2}, \dots, \be_N \},
\end{align}
where $M$ is the largest index such that $t_M \le \alpha + \Delta$. To aid subsequent notations, we define the index sets
\begin{align}
\cI_\alpha &= \{ 1, 2, \dots, M \},\\
\cI_\beta &= \{ M+1, M+2, \dots, N \}.
\end{align}
Define the temporal neighbours of each $\be_j \in \cE_\alpha$ as
\begin{align}
\cL_j = \{ k \in \cI_\beta \mid \left| t_k - t_j - \Delta \right| \le \epsilon_T \},
\end{align}
where $\epsilon_T$ is a user-determined threshold. Intuitively, $\cL_j$ is the subset of $\cE_\beta$ with a temporal gap of approximately $\Delta$ with $\be_j$. The tolerance of $\epsilon_T$ allows for time-stamping noise by the event sensor. Given a candidate $\bR_\Delta$, define
\begin{align}\label{eq:residual}
r_j(\bR_\Delta) = \min_{k \in \cL_j} \left\| \hat{\bu}_k - \bR_{\Delta} \hat{\bu}_j \right\|_2
\end{align}
as the \emph{residual} of event $\be_j$ from $\cE_\alpha$. The quantity
\begin{align}
\left\| \hat{\bu}_k - \bR_{\Delta} \hat{\bu}_j \right\|_2 \propto \angle(\hat{\bu}_k, \bR_{\Delta} \hat{\bu}_j)
\end{align}
measures the geometric misalignment between $\be_j$ and $\be_k \in \cL_j$. Computing $r_j(\bR_\Delta)$ implies searching for the ``best" spatiotemporally matching event from $\cE_\beta$ for $\be_j$ under $\bR_\delta$.

Our method simultaneously estimates $\bR_{\Delta}$ and event correspondences that are spatiotemporally consistent with $\bR_{\Delta}$ (up to temporal and geometric noise) by solving
\begin{align}\label{eq:str}
\min_{\bR_{\Delta}} \sum_{j = 1}^{K} r_{(j)}(\bR_\Delta),
\end{align}
where $K$ is a user-determined integer ($1 \le K \le M$), and $r_{(j)}(\bR_\Delta)$ is the $j$-th item of the ordered set
\begin{align}
\{ r_{(1)}(\bR_\Delta), r_{(2)}(\bR_\Delta), \dots, r_{(M)}(\bR_\Delta) \},
\end{align}
i.e., for all $j_1$ and $j_2$ such that $j_1 < j_2$,
\begin{align}
r_{(j_1)}(\bR_\Delta) \le r_{(j_2)}(\bR_\Delta).
\end{align}
Fundamentally, solving~\eqref{eq:str} finds the maximum likelihood 
estimate~\cite{besl1992method} of $\bR_\Delta$ from $\cE$. The usage of a ``trimming" parameter $K$ provides robustness against outliers~\cite{chetverikov2002trimmed}, i.e., events in $\cE_\alpha$ and $\cE_\beta$ without valid corresponding events.

Before describing the algorithm to solve~\eqref{eq:str}, let
\begin{align}
\tilde{\bR}_\Delta = \exp(\left[\tilde{\br}\right]_\times)
\end{align}
be the solution of~\eqref{eq:str}. Following the motion model~\eqref{eq:motionmodel}, we recover the angular velocity $\vel$ as
\begin{align}
\tilde{\vel} = \frac{2}{(\beta - \alpha)}\tilde{\br}.
\end{align}
Recall that our aim is to recover $\cM = \bR_{\alpha,\beta}$ from $\cE$. Referring to~\eqref{eq:motionmodel} again, we obtain
\begin{align}
\tilde{\bR}_{\alpha,\beta} = \exp(\left[2\tilde{\br}\right]_\times).
\end{align}
Sec.~\ref{sec:results} will investigate the performance of our approach.

\vspace{-1em}
\paragraph{Method}

Algorithm~\ref{alg:str} summarises a simple algorithm based on trimmed iterative closest points (TICP)~\cite{chetverikov2002trimmed,chetverikov2005robust} to solve~\eqref{eq:str} up to local optimality. Given an initial $\bR_\Delta = \bI$, the algorithm iterates two main steps to refine $\bR_\Delta$:
\begin{itemize}[topsep=0.25em,itemsep=0.25em,parsep=0em,leftmargin=1em]
\item Nearest neighbour search (Step~\ref{step:nn}), which produces a set of tentative event correspondences $\{ \langle \be_j, \be_{n_j} \rangle \}_{j=1}^M$ that are spatiotemporally consistent with the current $\bR_\Delta$.
\item Parameter update (Step~\ref{step:update}), which solves Wahba's problem~\cite{wahba1965least} on the $K$-most promising correspondences.
\end{itemize}
Given the short duration $\cT$, the angular separation of rays from events in $\cE_\alpha$ and $\cE_\beta$ are not significant and such a scheme is sufficient; as we will demonstrate in Sec.~\ref{sec:results}.

\begin{algorithm}[t]\centering
\caption{Spatiotemporal registration for event-based relative rotation estimation.}\label{alg:str}
\begin{algorithmic}[1]
\REQUIRE Event batch $\cE = \{ \be_i \}^{N}_{i=1} = \{(\bu_i, t_i, p_i)\}^{N}_{i=1}$ acquired over period $\cT = [\alpha,\beta]$, camera intrinsic matrix $\bK$, temporal threshold $\epsilon_T$, trimming parameter $K$.
\STATE $\Delta \leftarrow 0.5(\beta-\alpha)$.
\STATE $M \leftarrow \max_{i \in \{1,\dots,N\}}~i$ such that $t_i \le \alpha + \Delta$.
\STATE $\cI_\alpha \leftarrow \{1,\dots,M\}$.
\STATE $\cI_\beta \leftarrow \{M+1,\dots,N\}$.
\FOR{$j \in \cI_\alpha$}
\STATE $\cL_j \leftarrow \{ k \in \cI_\beta \mid \left| t_k - t_j - \Delta \right| \le \epsilon_T \}$.\label{step:tnn}
\ENDFOR
\STATE $\bR_{\Delta} \leftarrow \bI$
\WHILE{not converged}\label{step:iter1}
\FOR{$j \in \cI_\alpha$}
\STATE $n_j \leftarrow \argmin_{k \in \cL_j} \left\| \hat{\bu}_k - \bR_{\Delta} \hat{\bu}_j \right\|_2$.\label{step:nn}
\STATE $r_j \leftarrow \left\| \hat{\bu}_{n_j} - \bR_{\Delta} \hat{\bu}_{j} \right\|_2$.
\ENDFOR
\STATE $\{(1),\dots,(M)\} \leftarrow$ Index of sorting $\{r_1,\dots,r_M\}$.\label{step:ressort}
\STATE $\bR_\Delta \leftarrow \argmin_{\bR} \sum_{j=1}^K \left\| \hat{\bu}_{n_{(j)}} - \bR_{\Delta} \hat{\bu}_{(j)} \right\|_2$. \label{step:update}
\ENDWHILE\label{step:iter2}
\RETURN $\tilde{\bR}_\Delta = \bR_\Delta$ and $\{ \langle \be_{(j)}, \be_{n_{(j)}} \rangle \}^{K}_{j=1}$.
\end{algorithmic}
\end{algorithm}

To analyse Algorithm~\ref{alg:str}, we assume for simplicity
\begin{align}
|\cI_\alpha| = |\cI_\beta| = M = \frac{1}{2}N \equiv \cO(N).
\end{align}
A major task is to find the temporal neighbours in Step~\ref{step:tnn}. A naive approach is to compare each $t_j$ with $t_k$, which is $\cO(M^2)$. A more efficient technique is to index the intervals
\begin{align}
\left\{[t_k-\Delta,t_k+\Delta] \right\}^{}_{k \in \cI_\beta}
\end{align}
in an interval tree~\cite{de1997computational}, which takes $\cO(M \log M)$ time, then query the tree with each $t_j$ to find intervals that overlap with it in $\cO(\log M + m)$, where $m$ is the average size of $\cL_j$. Assuming events are distributed uniformly in $\cT$, we can take
\begin{align}
m \approx \frac{\Delta}{\epsilon_T}M.
\end{align}
By indexing each $\cL_j$ in a kd-tree, which takes time $\cO(m \log m)$, the nearest neighbour search in Step~\ref{step:nn} can be accomplished typically in time $\cO(\log m)$.  The remaining major operations are sorting the residuals (Step~\ref{step:ressort}), which can be done in $\cO(M\log M)$, and solving Wahba's problem (Step~\ref{step:update}), can be accomplished in $\cO(M)$ using singular value decomposition (SVD), which involves a matrix multiplication of $3\times M$ and $M\times 3$ and the $3\times 3$ matrix can be solve constantly.

The total cost of the Algorithm~\ref{alg:str} is thus
\begin{align}
\begin{aligned}
&\underbrace{\cO(M \log M)}_{\textrm{build interval tree}} + \underbrace{M \cO(\log M + m)}_{\textrm{query interval tree $M$ times}} + \underbrace{M\cO(m \log m)}_{\textrm{build $M$ kd-trees}} + \dots \\ \nonumber
&\underbrace{T( M\cO(\log m)  + \cO(M \log M)  + cO(M) )}_{\textrm{iterate Steps~\ref{step:iter1} to~\ref{step:iter2} $T$ times}}.
\end{aligned}
\end{align}
From our experiments, the algorithm typically takes $T = 10$ iterations to converge. Secs.~\ref{sec:results} and~\ref{sec:VOresults} will report the runtimes of our method. Note also that the cost of Algorithm~\ref{alg:str} does not depend on sensor resolution (number of pixels $P$).

\vspace{-1em}
\paragraph{Parameter setting}

The free parameters in Algorithm~\ref{alg:str} and their typical values are as follows:
\begin{itemize}[topsep=0.25em,itemsep=0.25em,parsep=0em,leftmargin=1em]
\item temporal threshold $\epsilon_T = 0.02(\beta-\alpha)$; and
\item trimming parameter $K = \lfloor 0.8M \rfloor$.
\end{itemize}

\subsection{RobotEvt dataset}\label{sec:dataset}

\begin{figure}[ht]\centering
	
	\includegraphics[width=0.8\columnwidth]{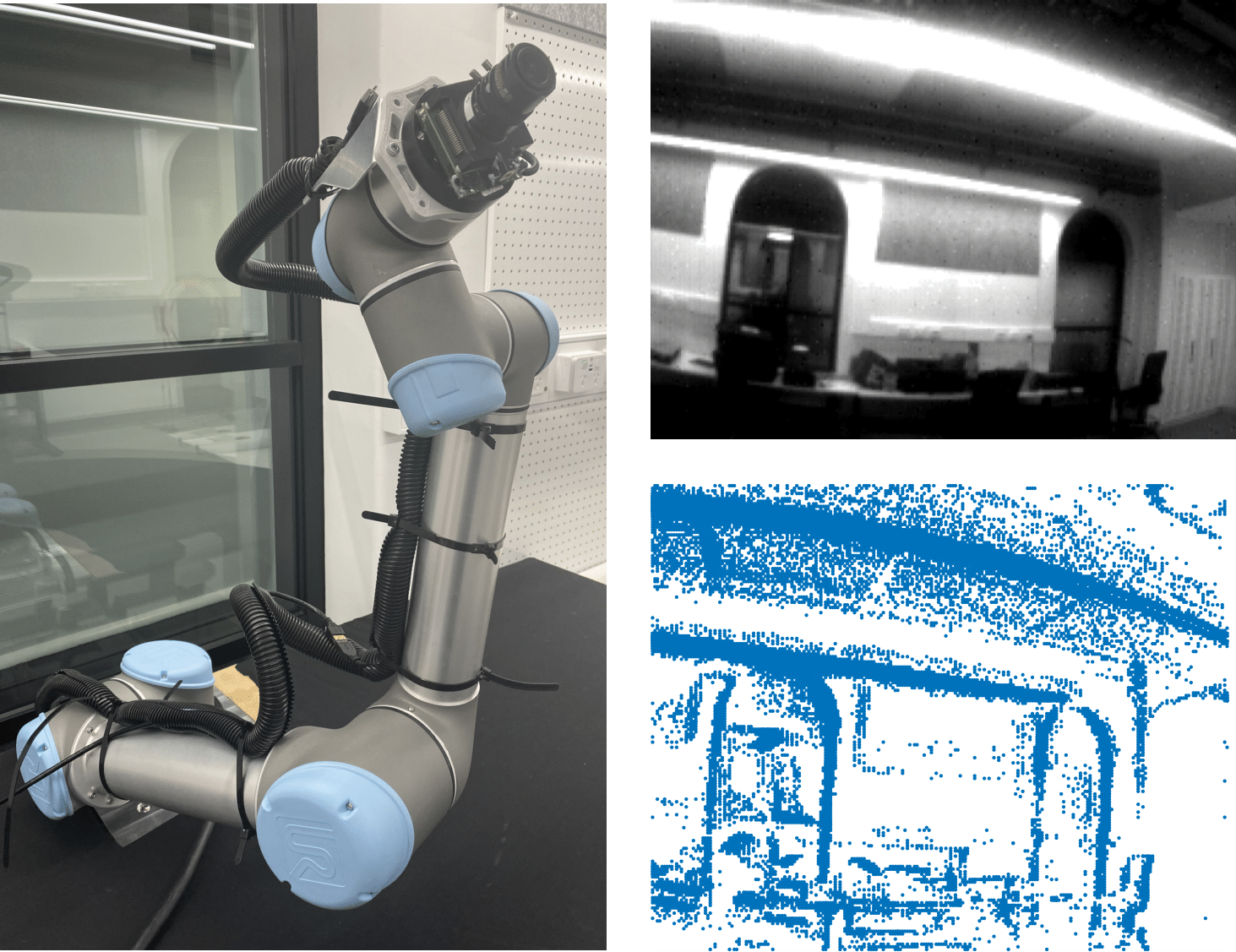}
	
	\caption{UR5 robot arm with DAVIS 240C event camera, and sample APS image and event image captured with our setup.}
	\label{fig:robot}
\end{figure}

To objectively evaluate STR, we construct an event dataset RobotEvt  using an iniVation DAVIS 240C event camera~\cite{davis240c} and a UR-5 robot arm~\cite{robot_arm_web}; see Fig.~\ref{fig:robot} for our setup. A number of event sequences were collected under different motion models, speeds and brightness from a static scene, specifically
\begin{itemize}[topsep=0.25em,itemsep=0.25em,parsep=0em,leftmargin=1em]
\item $4$ motion models: \texttt{PureRot} - pure rotation; \texttt{ParRot} - partial rotation; \texttt{PureTranslate} - pure translation and \texttt{FullMod} - full rigid motion model.
\item $3$ speeds: \texttt{Fast} - maximum speed of the robot arm ($1$ m/s); \texttt{Mid} - 75\% of the maximum speed and \texttt{Slow} - 50\% of the maximum speed.
\item $2$ brightness conditions: \texttt{On} and \texttt{Off} means bright and dark conditions, respectively.
\end{itemize}
In total, there are $4\times3\times2 = 24$ sequences, each of $60$~s duration and is named as a tuple of motion model, speed and brightness, e.g., \texttt{PureRot\_Fast\_On}.

Ground truth camera poses were extracted from the joint angles using the robot API in $125$ Hz. Radial undistortion for the event camera was also conducted prior to estimation.

\begin{figure}[ht]\centering
	\subfigure[Error vs batch duration $|\cT|$.]{\includegraphics[width=0.49\columnwidth]{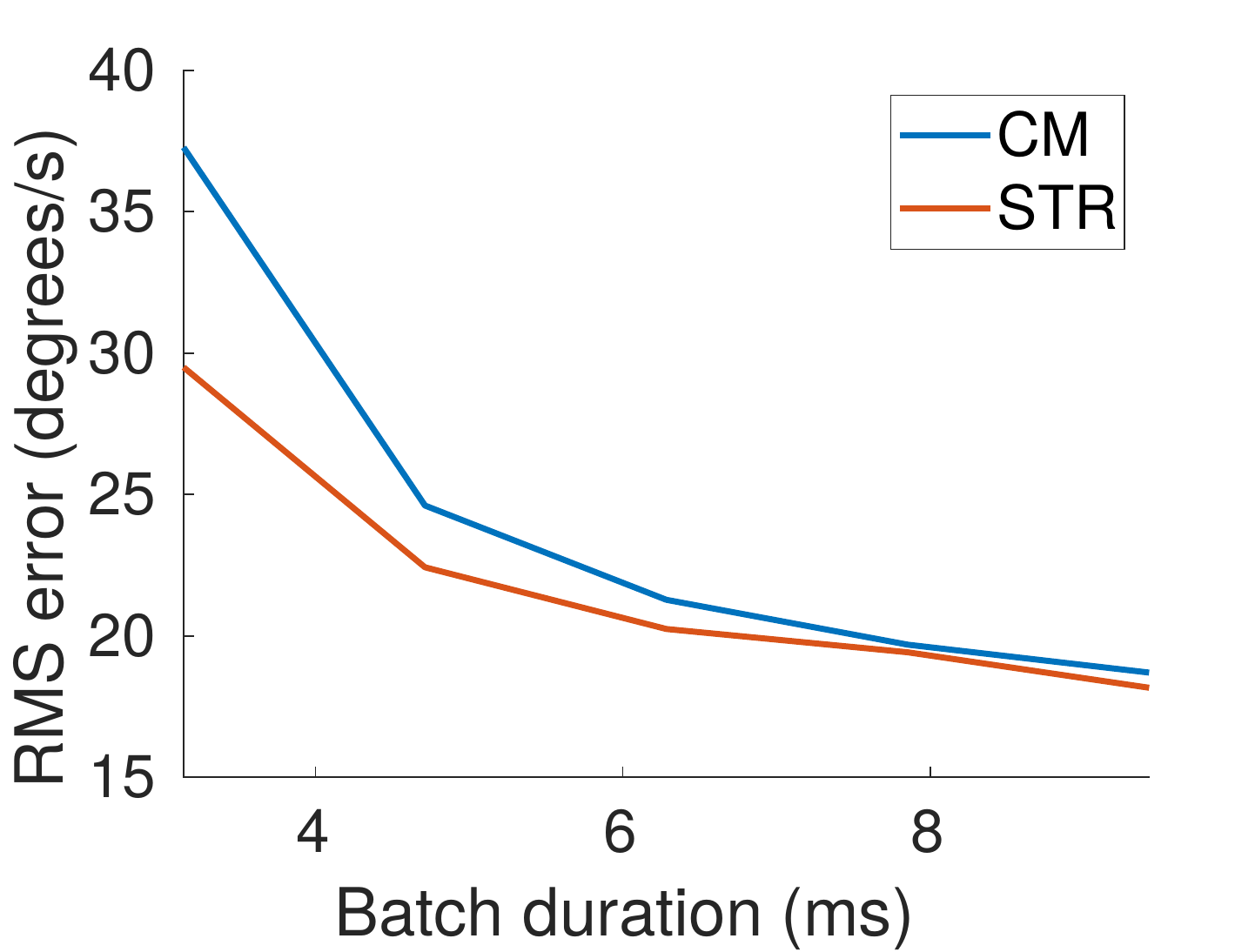}}
	\subfigure[Runtime vs batch size $N$.]{\includegraphics[width=0.49\columnwidth]{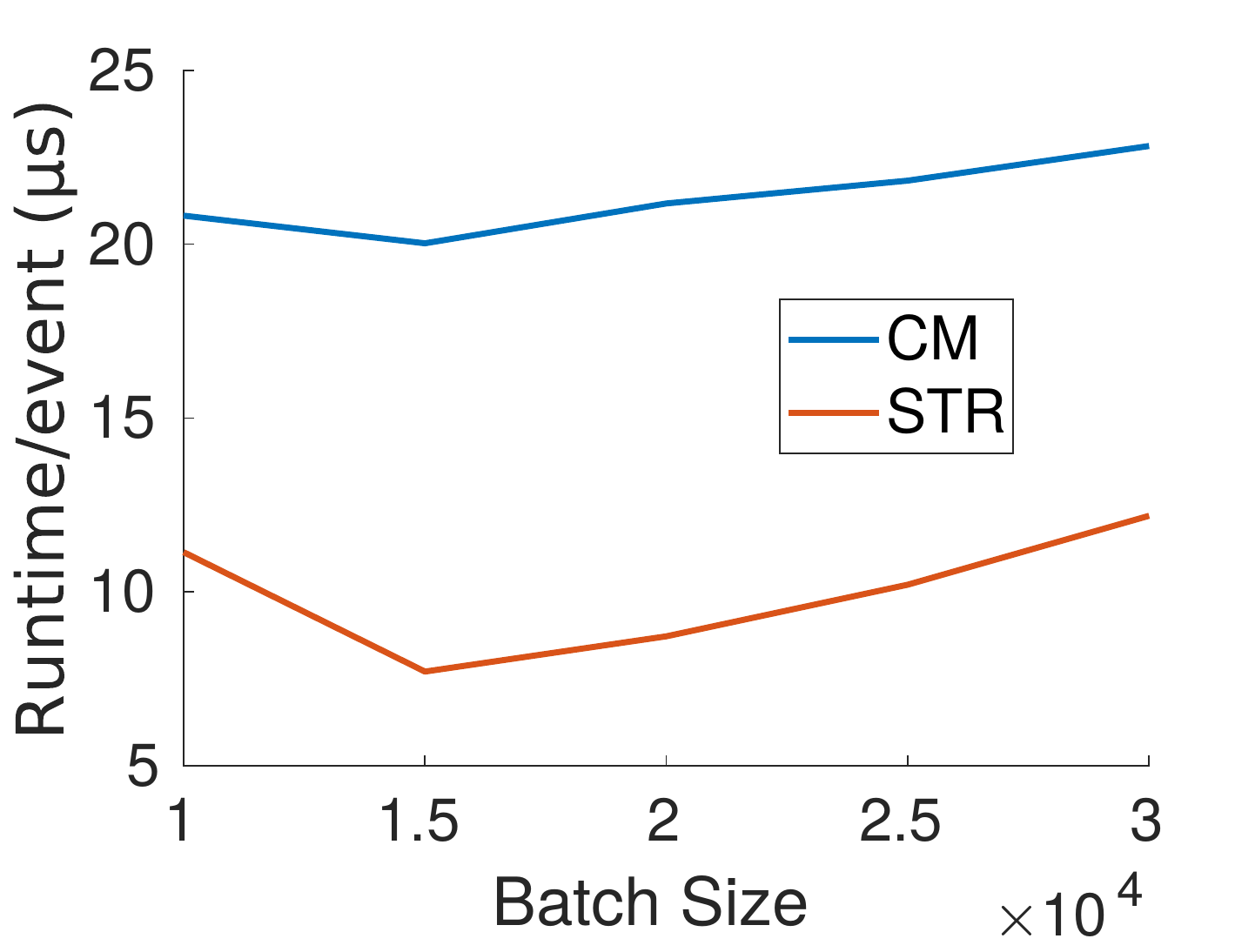}}
	\subfigure[Error vs batch duration $|\cT|$.]{\includegraphics[width=0.49\columnwidth]{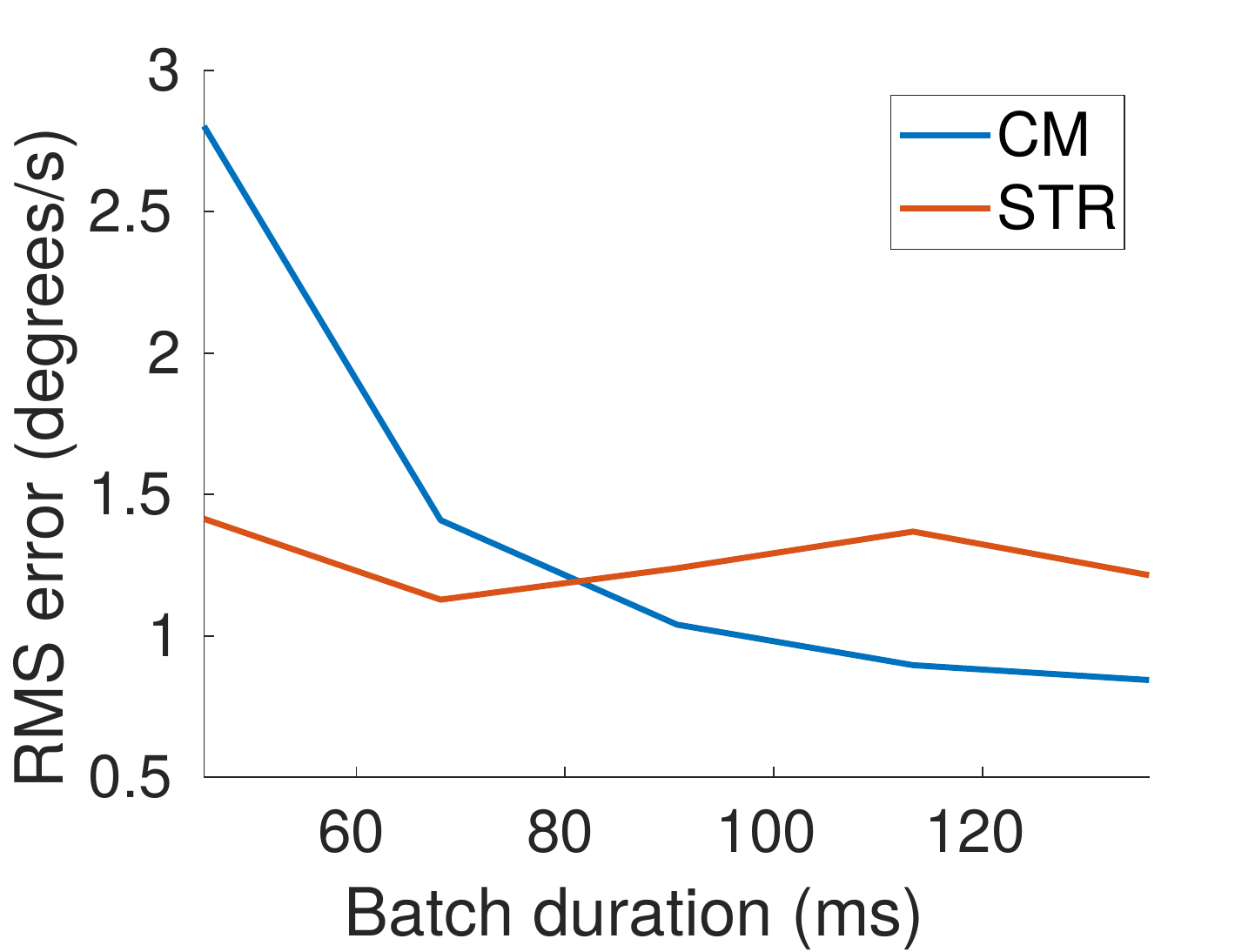}}
	\subfigure[Runtime vs batch size $N$.]{\includegraphics[width=0.49\columnwidth]{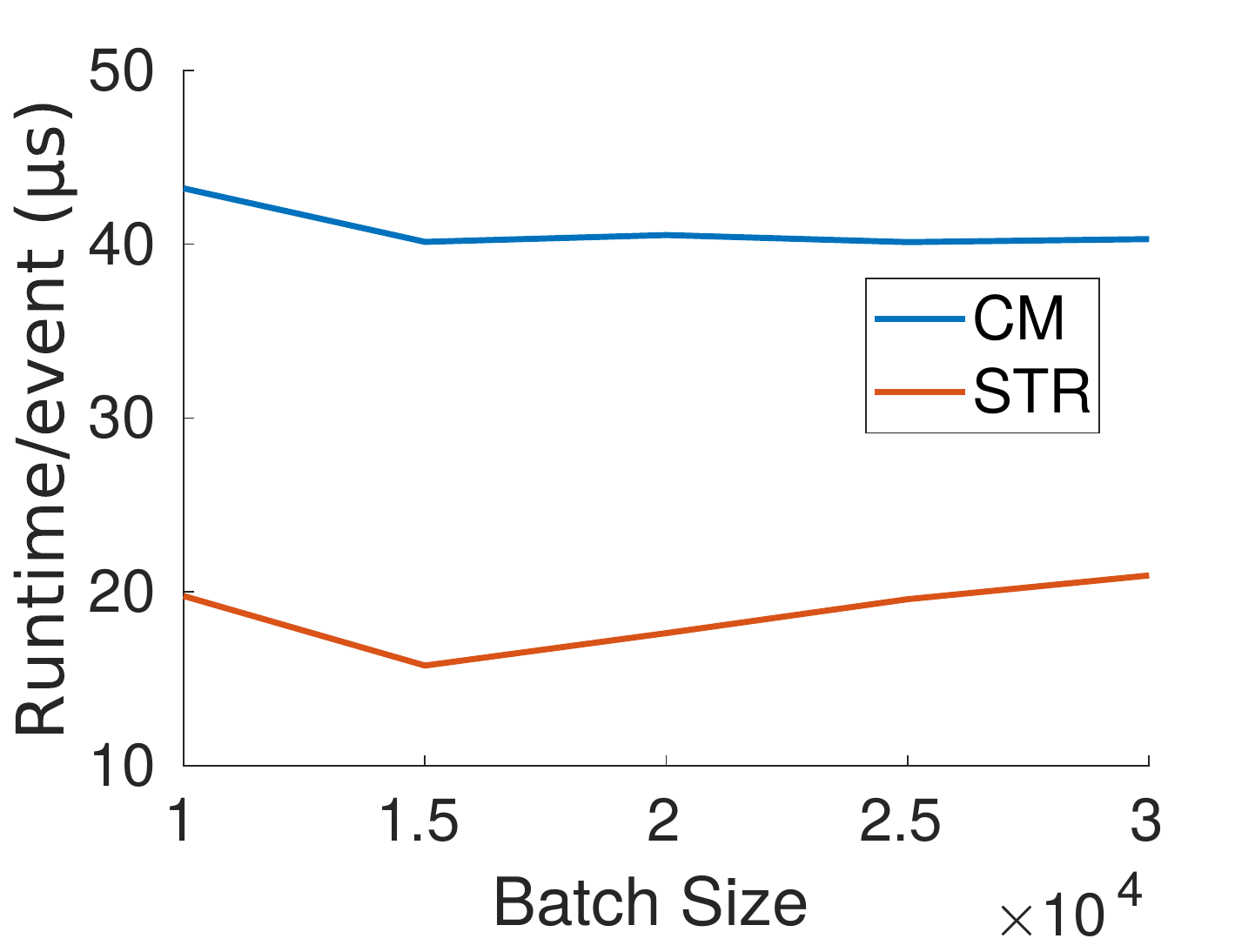}}
	\caption{RMS error (deg/s) and Runtime over batch size on \texttt{boxes} (top) and \texttt{PureRot\_Mid\_On} (bottom).}
	\label{fig:bath_varying}
\end{figure}

\subsection{Results}\label{sec:results}

We evaluate STR on the sequences with pure rotational motions from RobotEvt and the UZH dataset~\cite{ggler2017event} (specifically \texttt{poster}, \texttt{boxes}, \texttt{dynamic} and \texttt{shapes}). All sequences have 60 s duration with ground truth orientations.

From each sequence, we extract non-overlapping consecutive event batches $\cE$ of size $N = 10,000$ to $30,000$, estimate the relative rotation $\tilde{\bR}_\Delta$ from each batch and compared it against the ground truth $\bR^*_{\Delta}$ using
\begin{equation}\label{eq:angerr}
    d_{\angle}(\tilde{\bR}_\Delta,\bR^{*}_{\Delta}) = \|\log(\tilde{\bR}_\Delta{\bR^{*}_\Delta}^{T}) \|_2.
\end{equation}
The angular error~\eqref{eq:angerr} is then normalised by dividing with $\Delta$ to yield the angular velocity error (in deg/s).

We compared STR with CM~\cite{gallego2018unifying} and EM~\cite{Nunes_2020_ECCV}. All methods were implemented in C++ on a standard desktop with 3.0 GHz Intel i5 and 16 GB RAM. However, EM\footnote{\url{https://github.com/ImperialCollegeLondon/EventEMin}} was at least two orders of magnitude slower than STR; given the large number of batches to test (e.g., $> 7000$ batches in \texttt{Dynamic}), we leave the comparison with EM to Sec.~\ref{sec:VOresults}.

Figs.~\ref{fig:lowspeed} and~\ref{fig:bath_varying} show the RMS angular velocity error versus batch duration $|\cT|$ and runtime versus over batch size $N$ for \texttt{PureRot\_Mid\_Off}, \texttt{boxes} and \texttt{PureRot\_Mid\_On} (\textbf{see supplementary material} for more plots).  Table~\ref{tab:batch_size} records the statistics on the remaining sequences of Uth (with the exception of \texttt{Shapes} which we show in the supplementary material due to space constraints) and sequences of RobotEvt (\texttt{PureRot\_Off} indicates all pure rotational sequences in dark conditions; similarly for \texttt{PureRot\_On}).

The results show that as the batch size (or equivalently batch duration in this experiment) decreases, the accuracy of CM also decreases. This trend was more pronounced in RobotEvt, possibly due to the lower (but still substantial) speeds. In contrast, STR was able to maintain accuracy throughout the event batches, which indicates a higher temporal resolution than CM. Moreover, as demonstrated in Fig.~\ref{fig:runtime}, STR will not suffer from increasing resolution.

\begin{table}[]
	\centering
	{\footnotesize
	\begin{tabular}{|>{\centering\arraybackslash}p{0.82cm}|>{\centering\arraybackslash}p{0.95cm}>{\centering\arraybackslash}p{0.95cm}>{\centering\arraybackslash}p{0.95cm}>{\centering\arraybackslash}p{0.95cm}>{\centering\arraybackslash}p{0.95cm}|}
		\hline
		\multicolumn{6}{|c|}{\texttt{PureRot\_Off}}        \\
		\hline
		Method & $10000$ ($66$ms) & $15000$ ($99$ms) & $20000$ ($133$ms) & $25000$ ($166$ms) & $30000$ ($199$ms) \\
		\hline
		STR & $\bm{2.11}$ & $\bm{1.98}$    &   $\bm{1.91}$    & $\bm{1.91}$      &   $\bm{2.03}$         \\
		CM~\cite{gallego2018unifying} &  $3.93$    &   $2.24$    &   $1.98$    &  $1.96$     &  $2.22$    \\
		\hline
		\multicolumn{6}{|c|}{\texttt{PureRot\_On}}        \\
		\hline
		
		Method & $10000$ ($45$ms) & $15000$ ($68$ms) & $20000$ ($91$ms) & $25000$ ($113$ms) & $30000$ ($136$ms) \\
		\hline
		STR &   $\bm{2.42}$    &   $\bm{2.03}$    &   $\bm{1.91}$    &   $1.9$    &    $2.06$   \\
		CM~\cite{gallego2018unifying} &   $13.4$    &    $2.57$   &    $2.08$   &   $\bm{1.91}$    &   $\bm{1.88}$   \\
		\hline
		\multicolumn{6}{|c|}{\texttt{Dynamic}}        \\
		\hline
		Method & $10000$ ($8$ms) & $15000$ ($12$ms) & $20000$ ($16$ms) & $25000$ ($20$ms) & $30000$ ($24$ms) \\
		\hline
		STR &  $\bm{15.56}$     &   $\bm{13.46}$    &  $\bm{12.29}$   &   $\bm{11.69}$    & $11.33$      \\
		CM~\cite{gallego2018unifying} &   $17.46$    &   $14.24$    &   $12.91$    &   $11.72$    &  $\bm{11.23}$    \\
		\hline
		\multicolumn{6}{|c|}{\texttt{Poster}}        \\
		\hline
		Method & $10000$ ($3$ms) & $15000$ ($5$ms) & $20000$ ($7$ms) & $25000$ ($8$ms) & $30000$ ($10$ms)  \\
		\hline
		STR & $\bm{32.85}$      &    $\bm{28.36}$   &   $25.98$    &   $\bm{23.47}$    &   $\bm{22.30}$    \\
		CM~\cite{gallego2018unifying} &   $43.47$    &  $31.61$     &   $\bm{25.88}$    &  $24.57$     &   $23.03$   \\
		\hline
		
	\end{tabular}
	}
	\caption{RMS angular velocity error (deg/s) over all batches in pure rotation sequences in RobotEvt, \texttt{dynamic} and \texttt{poster}.}
	\label{tab:batch_size}
\end{table}

\subsection{Utilising depth information}

If depth information is available (e.g., by using stereo event cameras~\cite{zhu2018multivehicle}), we show how our method can be extended to estimate full rigid (6 DoF) motion.

Assuming constant angular velocity $\vel$ and linear velocity $\bv$ over $\cT$, the relative motion~\eqref{eq:relmotion} between $a,b \in \cT$ is
\begin{align}
\bM_{a,b} = \left[ \begin{matrix}  \bR_{a,b}  & -a\bR_{a,b}\bv_{a,b} + b\bv_{a,b}  \\ \bzero & 1 \end{matrix} \right].
\end{align}
Given two corresponding (and noiseless) events $\be = (\bu,d,t,p)$ and $\be^\prime = (\bu^\prime,d^\prime,t^\prime,p^\prime)$ in $\cT$, where $d$ and $d^\prime$ are respectively the depths of the events, the equation for geometric consistency in Definition~\ref{def} becomes
\begin{align}\label{eq:geom2}
d^\prime \hat{\bu}^\prime + t^\prime \bv_{a,b} = d\bR_{a,b}\hat{\bu} + t\bR_{a,b}\bv_{a,b}.
\end{align}
See supplementary material for the justification of~\eqref{eq:geom2}.

To estimate 6 DoF motion parameters ($\vel_\Delta, \bv_{\Delta}$) from a noisy event batch $\cE = \{ \be_i \}^{N}_{i=1} = \{(\bu_i,z_i,t_i, p_i)\}^{N}_{i=1}$ using Algorithm~\ref{alg:str}, we modify the residual~\eqref{eq:residual} to become
\begin{align*}
	\label{eq:residual6DoF}
	r_j(\bR_{\Delta},\bv_\Delta)=\min_{k \in \cL_j} \left\|d_k \hat{\bu}_k - d_j\bR_\Delta \hat{\bu}_j + t_j\bv_{\Delta} - t_j\bR_{a,b}\bv\right\|_2.
\end{align*}
The resulting update problem in (Step~\ref{step:update} in Algorithm~\ref{alg:str})
\begin{equation}\label{eq:str6DoF}
	\min_{\bR_{\Delta},\bv_\Delta} \sum_{j = 1}^{K} r_{(j)}(\bR_\Delta,\bv_\Delta)
\end{equation} 
can be solved using, e.g., gradient-based optimisation such as Levenberg Marquardt. We will leave 6 DoF event-based relative motion estimation as future work.

\section{Event-based visual odometry}\label{sec:evo}



A fundamental advantage of our relative rotation estimation method (Sec.~\ref{sec:str}) over previous techniques~\cite{gallego2018unifying,Nunes_2020_ECCV,Stoffregen_2019_CVPR,gallego2019focus} is that event correspondences are produced as a by-product, specifically by Step~\ref{step:nn} in Algorithm~\ref{alg:str}. We exploit this characteristic to track features across the event stream $\cS$ to build a rotational VO pipeline; see Algorithm~\ref{alg:evo}.

Given a fixed batch size $N$, Algorithm~\ref{alg:evo} accumulates overlapping event batches with ``stride" $0.5N$, i.e., if $\cE$ and $\cE^\prime$ are overlapping event batches, where
\begin{align}
\cT = [\alpha, \beta] \;\;\;\; \text{and} \;\;\;\; \cT^\prime = [\alpha^\prime , \beta^\prime]
\end{align}
are respectively the corresponding time windows, then $|\cE| = N$,  and $\cE^\prime = N$, and the batches have in common the set of events
\begin{align}\label{eq:overlap}
    \cF = \cE \cap \cE^\prime
\end{align}
in the time window $[\alpha^\prime,\beta]$, where $|\cF| = 0.5N$.

To conduct tracking, without loss of generality, let $\cE$ be the first batch. Executing Algorithm~\ref{alg:str} (STR) on $\cE$, we obtain the relative rotation and event correspondences
\begin{align}
\bR_{\alpha,\beta} \;\;\;\; \text{and} \;\;\;\; \{ \langle \be_j, \be_{n_{j}} \rangle \}^{K}_{j=1}.
\end{align}
Let $\bar{\cE}$ be the subset
\begin{align}
    \bar{\cE} = \{ \be_j, \be_{n_{j}} \}^{K}_{j=1} \cap \cF,
\end{align}
i.e., the subset of $\cE$ that contains only events that make up estimated correspondences that occurred in $[\alpha^\prime,\beta]$. Then, STR is performed on the reduced batch
\begin{align}
    \bar{\cE}^\prime = \bar{\cE} \cup \cE^\prime \setminus \cF
\end{align}
to estimate $\bR_{\alpha^\prime,\beta^\prime}$ and new event correspondences in $\cT^\prime$; see an illustration of the process in the supplementary material. By connecting the event correspondeces in $\cT$ and $\cT^\prime$, the process generates a set of $K$ event feature tracks
\begin{align}\label{eq:track}
\be_i \leftrightarrow \be_j \leftrightarrow \be_k
\end{align}
in the time window $[\alpha,\beta^\prime]$. By applying the same step on subsequent batches, the tracks can be extended (Step~\ref{step:ext_track} in Algorithm~\ref{alg:evo}). This obviates a separate feature detection and tracking heuristic~\cite{zhu2017event,alzugaray2018asynchronous,alzugaray2018ace,rebecq2017real}.


Different from Sec.~\ref{sec:str} we set the trimming parameter $K=\lfloor0.8|\bar{\cE}|\rfloor$, which decreases over batches. Since sufficient features tracks are necessary to perform STR, a "key batch" threshold $\epsilon_k$ is set to prevent the deficiency. If $K <\epsilon_k"$, the current batch $\cE'$ is designated a ey batch" and perform STR directly on $\cE'$ instead of $\bar{\cE}^\prime$ and reset $K=\lfloor0.4N\rfloor$. See Sec.~\ref{sec:VOresults} for concrete settings for $N$ and $\epsilon_k$.

Another crucial benefit of event feature tracking via Algorithm~\ref{alg:str} is enabling relative rotations to be computed between event batches, and allows the construction of a pose graph $\cG = (\cN,\cW)$, where the set of nodes
\begin{align}
\cN = \{ \cE^{(u)} \}^{U}_{u=1}
\end{align}
are event batches that share common tracked events observed thus far in the stream $\cS$. For any two $\cE^{(u)}$ and $\cE^{(v)}$ with common feature tracks, we solve~\eqref{eq:str} to get the relative rotation $\bR_{u,v}$ (see Step~\ref{step:estimate_rot} in Algorithm~\ref{alg:evo}). Given the relative rotations $\{\bR_{u,v}\}$, a robust rotation averaging problem~\cite{chatterjee2017robust}  
is solved to obtain the absolute orientations $\{ \bR_u \}$.


\begin{algorithm}[t]\centering
\caption{Event-based rotational VO with STR.}\label{alg:evo}
\begin{algorithmic}[1]
\REQUIRE Event stream $\cS$, camera intrinsic matrix $\bK$, temporal threshold $\epsilon_T$, batch size $N$ and "key batch" threshold $\epsilon_k$.
\STATE $ I \leftarrow 0$ and let $t_f$ be the duration of the event stream $\cS$.
\STATE $key \leftarrow true$, $\cN \leftarrow empty$.
\STATE $\cF\leftarrow\{(\bu_i,t_i,p_i)\}_{i=I+1}^{I+0.5N}$.
\WHILE{$t_{I+0.5N} < t_f$}
\STATE $I \leftarrow I+0.5N$ and $K\leftarrow 0.4N$.
\STATE $\cE_{\beta} = \{(\bu_i,t_i,p_i)\}_{i=I+1}^{I+0.5N}$.

\STATE \textbf{if} $key=true$  \textbf{then} $\cE \leftarrow \cF\cup \cE_{\beta}$ and $\cN\leftarrow\cF$.
\STATE \textbf{else} $\cE \leftarrow \bar{\cE} \cup \cE_{\beta}$ and $K\leftarrow 0.8|\bar{\cE}|$.
\STATE $key\leftarrow false$ and $\cF\leftarrow \cE_{\beta}$.
\STATE $\bR_{\Delta},\{ \langle \be_j, \be_{n_j} \rangle \}_{j=1}^K\leftarrow$ STR($\cE$,$\bK$,$\epsilon_T$,$K$).
\STATE $\bar{\cE} = \{ \be_j, \be_{n_{j}} \}^{K}_{j=1} \cap \cF$.
\STATE $\cN \leftarrow \cN \cup \bar{\cE}$. \label{step:ext_track}
\STATE \textbf{if} $K>\epsilon_k$ \textbf{then} continue.\label{step:check_sufficient}
\STATE 
 $\{\bR_{u,v}\}_{u,v\in U} \leftarrow Est\_Rot(\cE_u,\cE_v)$\label{step:estimate_rot}.
 \STATE $\{\bR_u\}_{u\in U}\leftarrow Rot\_Avg(\{\bR_{u,v}\}_{u,v\in U})$.
 \STATE $\cN \leftarrow empty$ and $key \leftarrow true$.\label{step:rot_avg}
\ENDWHILE
\RETURN $\{\bR_u\}$.
\end{algorithmic}
\end{algorithm}




\subsection{VO results}\label{sec:VOresults}

We benchmarked our VO technique against the following approaches on the datasets employed in Sec.~\ref{sec:results}:
\begin{itemize}[topsep=0.25em,itemsep=0.25em,parsep=0em,leftmargin=1em]
	\item VCM~\cite{gallego2018unifying}: absolute orientations were computed by chaining relative rotations from CM.
	\item VEM~\cite{Nunes_2020_ECCV}: entropy maximisation method. Absolute orientations were generated by chaining.
	\item ZHU~\cite{zhu2017event}: probabilistic feature tracking method. It's an indirect approach (different to CM, EM and our STR), which extract and track features frrm event stream. We used the feature tracks to calculate the relative rotations, and absolute orientations were generated by chaining. For fair comparisons, we disabled the IMU input to ZHU.
\end{itemize}
We also tested our method with and without rotation averaging (VSTR$_\text{A}$ and VSTR$_\text{C}$). All methods operated on event batches of size $N = 30,000$ for all sequences and $\epsilon_k = 2,000$ for VSTR$_\text{C}$ and VSTR$_\text{A}$.

\begin{figure}
\centering
\subfigure{\includegraphics[width=0.99\linewidth]{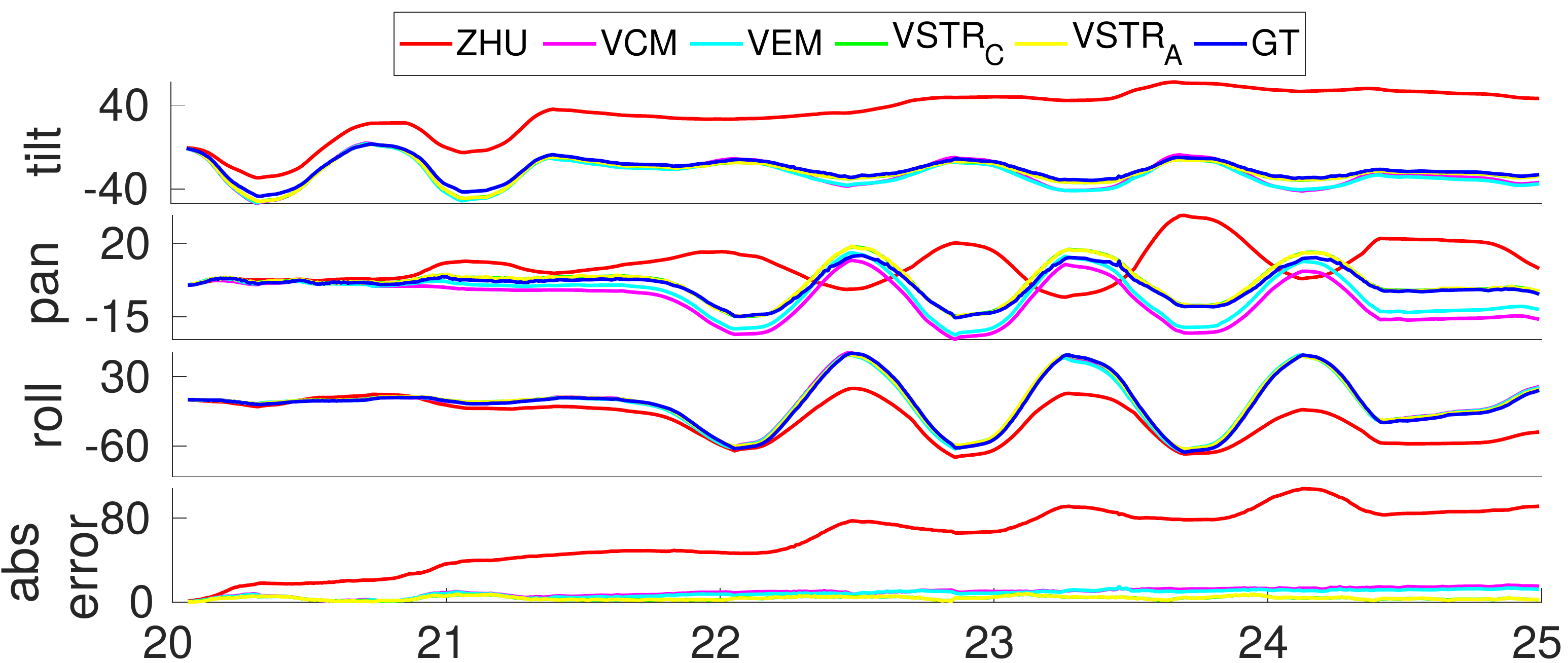}}\\
\subfigure{\includegraphics[width=0.99\linewidth]{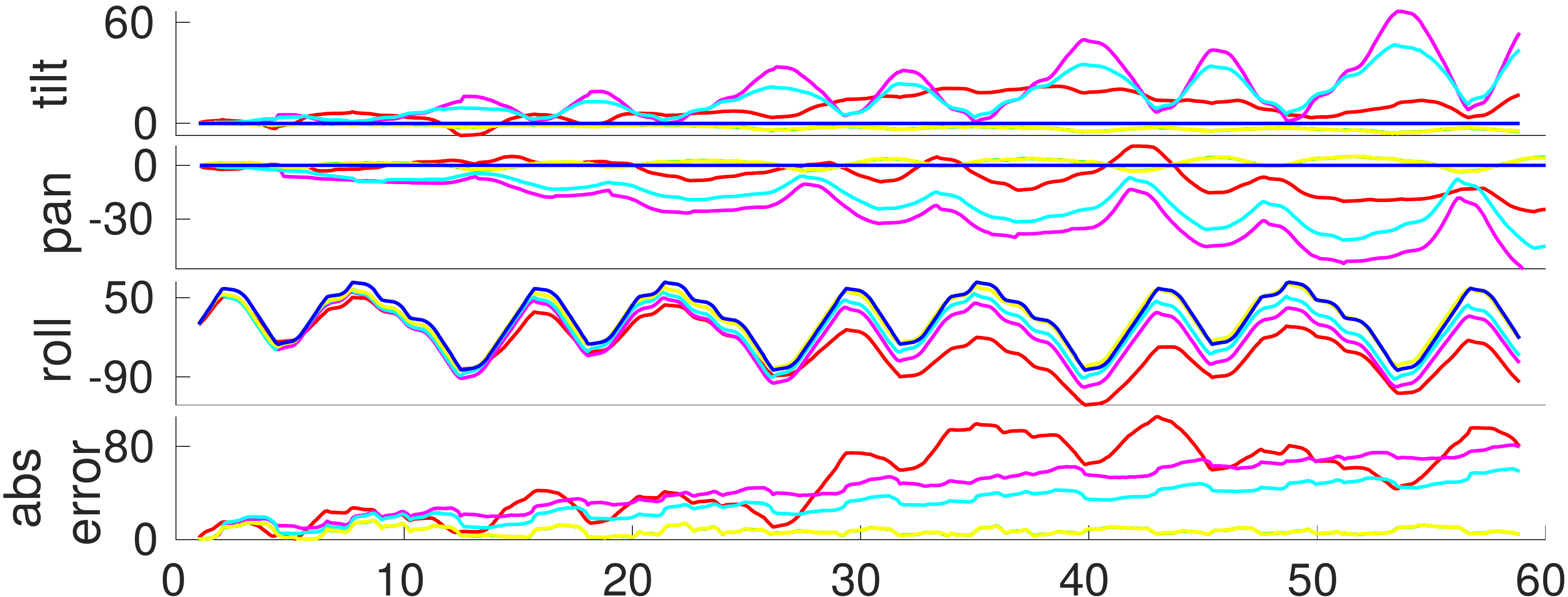}}
\caption{Absolute orientation trajectories (plotted as Euler angles) and absolute orientation error over time. Top: \texttt{Poster}. Bottom: \texttt{PureRot\_Fast\_On}.}
\label{fig:vo_quan}
\end{figure}

Fig.~\ref{fig:vo_quan} plots the absolute orientation trajectories and absolute orientation error for \texttt{boxes} and \texttt{PureRot\_Fast\_On}, where the length of the graphs are optimised for visualisation (\textbf{see supplementary material} for full results). Table~\ref{tab:abs_err} depicts the average absolute orientation error (in deg) of each 60 s sequence and average runtime of the pipelines. VSTR achieved the best accuracy amongst most sequences and was the fastest (Note that our STR can process $1,000,000$ events/s when $N=20,000$ without sacrificing accuracy.). VEM achieved comparable accuracy but was much slower than VSTR, while VCM was slightly worse in accuracy and runtime than VSTR. The high error of ZHU indicated dependence on the IMU for tracking.

\begin{table}[]
	\centering
	{\footnotesize
	\begin{tabular}{|>{\centering\arraybackslash}p{1.9cm}|>{\centering\arraybackslash}p{0.7cm}>{\centering\arraybackslash}p{0.7cm}>{\centering\arraybackslash}p{0.7cm}>{\centering\arraybackslash}p{0.7cm}>{\centering\arraybackslash}p{1.1cm}|}
		\hline
		Sequences   & VSTR$_\text{C}$ & VSTR$_\text{A}$ & VCM  & VEM & ZHU \\
		\hline
		\texttt{PureRot\_On} &   $\bm{5.11}$              &   $5.13$            &  $34.46$     &   $16.64$     &   $44.54$      \\
		\texttt{PureRot\_Off}&    $6.05$             &     $\bm{6.01}$         &  $29.38$    &   $7.53$    &  $46.76$     \\
		\texttt{boxes}       &        $11.40$         &      $\bm{11.38}$         &   $24.82$    &   $26.16$     &  $143.55$       \\
		\texttt{dynamic}       &    $21.75$             &     $21.78$          &  $26.68$      &    $\bm{6.36}$     &    $125.69$      \\
		\texttt{poster}      &        $12.97$         &       $\bm{12.06}$        &    $49.75$   &  $45.86$      &  $132.69$   \\
		\hline   
		\hline
		Runtime (s) & $\bm{0.061}$ & $0.062$ & $0.538$ & $2.313$ & $2.863$ \\ 
		\hline
	\end{tabular}
	\caption{Average absolute orientation error (deg) and average runtime per batch over all instances in \texttt{PureRot}, \texttt{boxes}, \texttt{dynamic} and \texttt{poster} sequences.}
	\label{tab:abs_err}
}
\end{table}
\section{Conclusions}

Accurate event-based motion estimation can be accomplished using much simpler techniques than CM, EM with less computational resources. The theoretical justification of our STR has been conducted, and the experiments showed that fewer parameters tunning are needed for different scenarios. Furthermore, the feature tracks generated by the our STR can be incorporated in solving loop closure.

\section*{Acknowledgement}
This work was supported by Australian Research Council ARC DP200101675.
\vfill

\pagebreak

{\small
\bibliographystyle{ieee_fullname}
\bibliography{spatiotemporal}
}

\end{document}